\pdfoutput=1

\documentclass[11pt]{article}

\usepackage{acl}
\usepackage{times}
\usepackage{latexsym}

\usepackage[T1]{fontenc}

\usepackage[utf8]{inputenc}
\usepackage{microtype}
\usepackage{booktabs}
\usepackage{tabularx}
\usepackage{multirow}
\usepackage{multicol}

\usepackage{float}
\usepackage{amsmath}
\usepackage{amsfonts}
\usepackage{verbatim}
\usepackage{graphicx}
\usepackage{relsize}
\usepackage{url}
\usepackage{xspace}
\usepackage{cleveref}
\usepackage{adjustbox}

\usepackage{xcolor}
\usepackage{tcolorbox}

\newcommand{\tabref}[1]{Table~\ref{#1}\xspace}
\newcommand{\figref}[1]{Figure~\ref{#1}\xspace}

\newcommand{\secref}[1]{Section\xspace\ref{#1}\xspace}
\newcommand{\appref}[1]{Appendix\xspace\ref{#1}\xspace}
\newcommand{\myparagraph}[1]{\noindent\textbf{#1}\xspace}

\newcommand{\z}{\phantom{0}}

\newcommand{\lamb}{\textsc{Lamb}\xspace}
\newcommand{\selsum}{\textsc{SelSum}\xspace}
\newcommand{\amasum}{\textsc{AmaSum}\xspace}
\newcommand{\ex}[1]{\textit{#1}\xspace}
\newcommand{\token}[1]{\texttt{#1}\xspace}

\title{Location Aware Modular Biencoder for Tourism Question Answering}

\author{Haonan Li$^{\spadesuit,\clubsuit}$  \hspace{1cm} Martin Tomko$^\heartsuit$ \hspace{1cm} Timothy Baldwin$^{\spadesuit,\clubsuit}$\\
	$\spadesuit$ School of Computing and Information Systems, The University of Melbourne  \\
	$\heartsuit$ Department of Infrastructure Engineering, The University of Melbourne \\
	$\clubsuit$ Department of Natural Language Processing, MBZUAI \\
	\tt haonan.li@mbzuai.ac.ae,  
	\tt tomkom@unimelb.edu.au, tb@ldwin.net\\
}

\begin{document}
\maketitle
\begin{abstract}
Answering real-world tourism questions that seek Point-of-Interest (POI) recommendations is challenging, as it requires both spatial and non-spatial reasoning, over a large candidate pool.
The traditional method of encoding each pair of question and POI becomes inefficient when the number of candidates increases, making it infeasible for real-world applications.
To overcome this, we propose treating the QA task as a dense vector retrieval problem, where we encode questions and POIs separately and retrieve the most relevant POIs for a question by utilizing embedding space similarity. 
We use pretrained language models (PLMs) to encode textual information, and train a location encoder to capture spatial information of POIs.
Experiments on a real-world tourism QA dataset demonstrate that our approach is effective, efficient, and outperforms previous methods across all metrics.
Enabled by the dense retrieval architecture, we further build a global evaluation baseline, expanding the search space by 20 times compared to previous work.
We also explore several factors that impact on the model's performance through follow-up experiments.
Our code and model are publicly available at \url{https://github.com/haonan-li/LAMB}.
\end{abstract}

\section{Introduction}

Question answering (QA) models and recommender systems have undergone rapid development in recent years \cite{bidaf,SQuAD,NQ,lee2019latent,cui2020personalized,hamid2021smart}.
However, personalised question answering is still highly challenging and relatively unexplored in the literature.
Consider the example question in \figref{fig:example}, in the form of a real-world point-of-interest (POI) recommendation question from a travel forum.
Answering such questions requires understanding of the question text with possibly explicit (e.g., \ex{in Dublin}) or vague and ambiguous (e.g., \ex{within walking distance of Grafton Street}) spatial constraints,
as well as a fast indexing method that supports large-scale reasoning over both spatial and non-spatial (e.g., \ex{fairly priced restaurants}) constraints.
\begin{figure}[t]
	\centering
	\begin{tcolorbox}[colback=white,arc=0mm,left=0mm,right=0mm,top=1mm,bottom=1mm]
		\textbf{Question:} \ex{Hi! My wife and I are in our late thirties and going to be \textcolor{brown}{in Dublin} on September 28 and 29th. We are staying in the Grafton Street area. Does anybody have any suggestions for some \textcolor{olive}{fairly priced restaurants} with \textcolor{olive}{great food} \textcolor{brown}{within walking distance of Grafton Street}? Also, What about some \textcolor{olive}{good pubs with live local music}? (I realize it is a Sun and Mon night and may be slow) Any suggestion would be appreciated! Thanks!}\\
		\textbf{Answer ID}: \ex{11\_R\_4392} \\
		\textbf{Answer Name}: The Porterhouse Central, 45-47 Nassau Street, Dublin.
	\end{tcolorbox}
	\caption{An example of real-world POI recommendation question from the TourismQA dataset \cite{contractor2019large}. Colored text represents constraints relevant to recommending POIs.}
	\label{fig:example}
\end{figure}

Recently, there has been increased interest in geospatial QA. 
Most approaches focus on querying structured knowledge bases, based on translating natural language questions into structured queries, e.g., using SPARQL \cite{punjani2018template,li2021neural,DBLP:conf/www/HamzeiT022}. 
Separately, \citet{contractor2019large} introduced the task of answering POI-seeking questions using geospatial metadata and reviews that describe POIs.
In later work, they proposed a spatial--textual reasoning network that uses distance-aware question embeddings as input and encodes question--POI pairs using attention \citep{contractor2021joint}.
However, as their model creates separate question embeddings for each POI, the inference cost increases linearly in the number of POIs, and the model is incompatible with large pre-trained models such as BERT \cite{BERT} or even medium-sized QA models such as BiDAF \cite{bidaf}.

In this work, we address the question: can we build a more efficient POI recommendation system which supports the use of advanced pre-trained language models as the textual encoder?
By presenting the {\bf L}ocation {\bf a}ware {\bf m}odular {\bf b}i-encoder (``\lamb'') model.
We use a bi-encoder architecture to encode questions and POIs separately, where the question encoder is a textual module and the POI encoder consists of a textual and a location module.
By encoding them separately, we cast the task as a retrieval problem based on dense vector similarity between the question and each POI.
For training, we combine each question with one positively-labeled POI and multiple negatively-labeled POIs, and use contrastive learning to train the question encoder and POI encoder simultaneously, by maximizing the similarity between the question and positive POI.
After training, we generate location-aware dense representations for all POIs using the POI encoder, and index them by city name and entity (POI) type.
For inference, we use the question encoder to generate a location-aware representation, and rank the POIs using similarity.

Our contributions are four-fold:  
(1) we propose a location-aware modular bi-encoder model which fuses spatial and textual information;
(2) we demonstrate that the proposed model outperforms the existing SOTA on a real-world tourism QA dataset, with huge improvements in training and inference efficiency; 
(3) we build new global evaluation baselines by expanding the search space 20$\times$ over local evaluation; and finally, 
(4) we analyse the influence of different training strategies and hyper-parameters through extensive experiments.

\begin{figure*}[t]
	\includegraphics[width=1\textwidth]{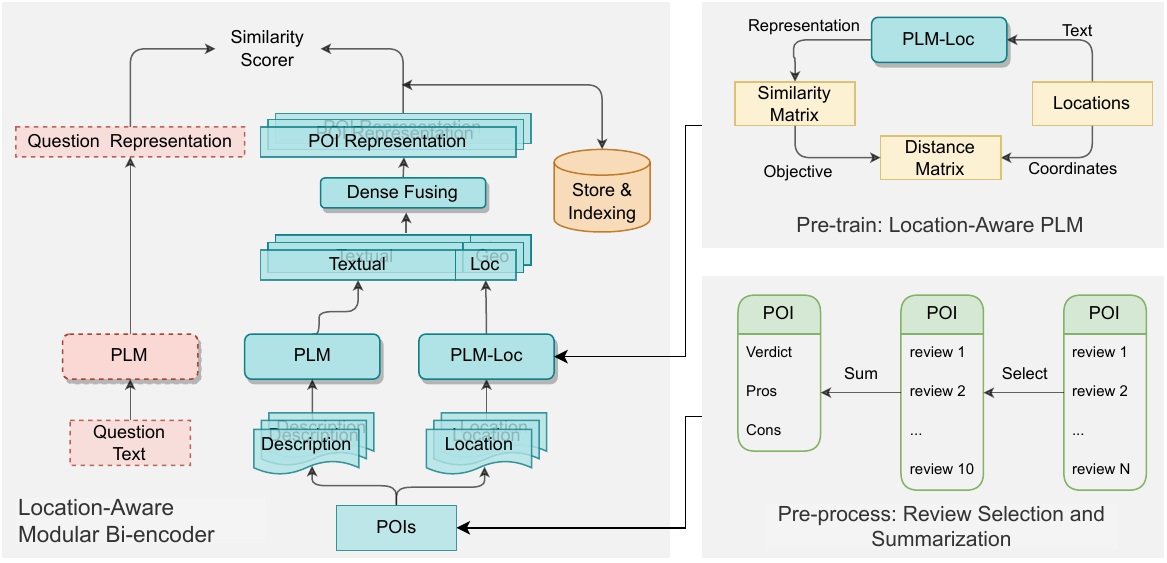}
    \caption{Proposed approach. The reviews of POIs are first selected and summarized by \selsum (bottom right part). A location module is separately pre-trained under the supervision of geocoordinate-based distances (top right part).
    The left part is the main \lamb model. The cyan- and salmon-coloured parts are the POI encoder and question encoder, respectively. The orange part is the index of POI embeddings used for inference.}
	\label{fig:model}
\end{figure*}

\section{Methodology}

In this section, we first formulate the task, and then introduce the POI pre-processing method and the \lamb model. 
Finally, we describe the efficient training and inference strategies.

\subsection{Task Formulation}

Given a question $q$, the task is to find the most probable POI answer $p$ from a candidate pool $P$, which satisfies spatial and non-spatial constraints in $q$.
Each POI in $P$ consists of a \ex{geo-coordinates} $(lat, long)$ of the POI, the multi-granularity location \ex{name} (POI entity name, street, city, postcode), and a list of textual \ex{reviews} $ = (r_1, r_2, ... r_n)$. 
It can be represented as $p = \langle\ex{coordinates}, \ex{name}, \ex{reviews}\rangle$ (see \appref{sec:poi} for an example).

\subsection{POI Pre-processing}\label{sec:selsum}

Reviews of POIs provide useful information to represent POIs, however, each candidate can have hundreds of reviews, the total length greatly exceeding the maximum token length of 512 tokens in general PLMs such as BERT.
To choose more representative reviews, previous work \cite{contractor2019large} has clustered reviews into $K$ clusters, and then represented the POI using the top-$N$ sentences from each cluster based on distance from the cluster centroid, resulting in $N\times K$ sentences.
However, this approach is potentially problematic as clusters can be of varying size and density, and outliers can affect the centroid. 
To keep representative reviews, $K$ and $N$ should not be too small, e.g., \citet{contractor2019large,contractor2021joint} set $N=K=10$.

In this paper, we adopt the \selsum \citep{SelSum} model, which consists of a selector to choose the $M$ most representative reviews and a summarizer to generate a summary of the selected reviews. 
We use a model pre-trained on the \amasum dataset, which includes verdicts, pros, and cons, and hundreds of reviews for more than 31,000 summarized Amazon products (see example in \appref{sec:selsum_app}).
We compare the results using clustering, the selection module only, and the full \selsum model in \appref{sec:selsum_app}. 
Our results show that using a 3-sentence summary for each POI achieves comparable results with a clustering approach that represents each POI via 100 sentences, and that using 10 sentences outperforms the clustering method.

\subsection{Location Aware Modular Bi-encoder}\label{sec:lamb}
\lamb (see \figref{fig:model}) uses a bi-encoder framework to encode questions and POIs. 
The question encoder is a textual module which takes question text as input, and outputs dense representations.
The POI encoder consists of a textual module and a location module, where the textual module encodes a description and/or reviews associated with it, and the location module encodes the multi-granularity location names.
The outputs of the textual and location modules are real-valued vectors, which are concatenated to represent a POI.
Full details of the model are presented below.

\paragraph{Textual Module}
We use two independent PLMs as the textual encoder for questions and POIs, using the \token{[CLS]} token representation as the output. 
For questions, we do not preprocess the question text, while for POIs, we concatenate the preprocessed reviews.

\paragraph{Location Module}\label{sec:location}
Spatial constraints are crucial in retrieving relevant POIs to a question. 
However, previous research has shown that PLMs perform poorly in encoding and reasoning over spatial data, especially for geolocation information \citep{scherrer2021social,hofmann2022geographic}.
To enhance the model's ability to capture geospatial information, we employ a location module that explicitly encodes the multi-granularity location name of a POI into a dense vector.
We initialize the location module by choosing several transformer blocks from a PLM, and continue pre-training it to learn geo-coordinate-aware location name representations.
The training object is designed to pull together pairs of encoded location representations if the locations are physically near each other, and push them apart if they are far from each other.

Formally, for any three POIs $(p_0, p_1, p_2)$, suppose the corresponding locations are $(l_0, l_1, l_2)$, and the encoded representations are $(h_0, h_1, h_2)$.
Here $l_i (i=0,1,2)$ is a 1-d vector $[lat_i, long_i]$, representing the latitude and longitude of $p_i$, with $lat_i \in [-90, 90]$ and $long_i \in [-180, 180]$, and $h_i$ is a vector.
We choose $p_0$ to be an anchor location, and $d_i (i=1,2) \in [0,1]$ to represent the normalized Haversine distance between $l_0$ and $l_i$, representing the greater-circle distance between two points on a sphere. 
Similarly, $s_i (i=1,2) \in [0,1]$ represents the cosine similarity between $h_0$ and $h_i$. 
We use the triplet margin loss, and define the loss function as follows:
\begin{equation*}
    \small
    \mathcal{L} = \left\{\begin{matrix}
            \max((s_1 - s_2) + (d_1 - d_2), 0)         & \text{if}  (d_1 - d_2) > 0 \\
        \max((s_2 - s_1) - (d_1 - d_2), 0)         & \text{otherwise} \\
    \end{matrix}\right.
\end{equation*}
In the first case, $d_1-d_2>0$ means that $p_2$ is closer to $p_0$ than $p_1$, and hence we structure the loss to learn a larger $s_2$ (= higher similarity between $p_0$ and $p_2$) and smaller $s_1$ (= lower similarity between $p_0$ and $p_1$).
We set the difference between the two distances as a dynamic margin, which controls the rationally-valued similarity difference.

\paragraph{Question and POI Encoders}
As mentioned above, we use a separate textual encoding module $E_{P}^{text}$ and location encoding module $E_{P}^{loc}$ to encode each POI. 
These modules map the review text and location names to fixed-length vectors:
\begin{align*}
    r_{p}^{text} &= E_{P}^{text}(p) \in \mathbb{R}^{1\times d_1} \\
    r_{p}^{loc} &= E_{P}^{loc}(p) \in \mathbb{R}^{1\times d_2} 
\end{align*}
We concatenate $r_{p}^{text}$ and $r_{p}^{loc}$ and then use a dense layer to fuse the representations together, resulting in the POI representation $r_p \in \mathbb{R}^{1\times d}$:
\begin{equation*}
    r_p = \text{Dense}([r_{p}^{text}, r_{p}^{loc}]) \in \mathbb{R}^{1\times d}
\end{equation*}
For questions, we similarly tried using separate text and location modules, and combining their outputs.
However, we found that the text may contain distractor locations that should not be considered as spatial constraints, and that context is essential.
(e.g., the place name \ex{Italy} in question \ex{Hey I am from Italy, please suggest a restaurant in Berlin that suits my appetite.})
Hence, we use a single textual module $E_{Q}^{text}$ which directly maps the question text into representation $r_q \in \mathbb{R}^{1\times d}$, of the same dimension as a POI.

\subsection{Training and Inference}\label{sec:training}

We train the two encoders simultaneously using contrastive learning. 
We input each question $q_i$ with one positive POI $p_i^{+}$ and several negative POIs $p_{i,1}^{-},...p_{i,n}^{-}$ into the model,
with the objective to maximize the similarity between the embeddings of $q_i$ and $p_i^{+}$, while minimizing the similarity between the embeddings of $q_i$ and $p_{i,1}^{-},...p_{i,n}^{-}$.
We use the negative log-likelihood (NLL) loss of the positive POIs as our objective function:
\begin{align*}
&\mathcal{L}(q_{i},p_{i}^{+},p_{i,1}^{-},...p_{i,n}^{-})  \nonumber \\
= &- \log \frac{e^{\text{sim}(q_i,p_i^{+})}}{e^{\text{sim}(q_i,p_i^{+})}+\sum_{j=1}^{n} e^{\text{sim}(q_i, p_{i,j}^{-})}} 
\end{align*}
where similarity function $\text{sim}(p, q)$ is the inner product.

\myparagraph{Negative Sampling Strategy}
A critical question in contrastive learning is how to construct positive and negative examples. In our case, for each question, there can be more than one answer (= positive) POI. 
To make use of every positive POI, as well as to adapt to the NLL loss function, we create a training example for each positive POI.
For negative samples, all non-answer POIs are candidate negative samples, but previous work \cite{DPR,DBLP:conf/iclr/XiongXLTLBAO21} has shown that high-quality negative samples help to learn a better encoder.
In this research, we consider three different types of negative samples: (1) \ex{easy negatives} = random (non-answer) POIs from the entire candidate set; (2) \ex{medium negatives} = random (non-answer) POIs that are in the same city and of the same type (restaurant, attraction, or hotel) as the answer POI; and (3) \ex{hard negatives} = top-$k$ ranked non-answer POIs from the previous epoch.

\myparagraph{Two-phase Training}
We conduct two phases of training: first, we use easy and medium negatives to do warm-up training of the model, and provide the model with a relatively easily-optimizable objective; next, we switch over to training with a mixture of medium and hard negatives.\footnote{For convenience, we use ``easy'' and ``hard'' negatives to describe the training setting in any single phase.}
We sample hard negatives by performing inference on the training data after each epoch (or a specific number of steps) to find the top-$k$ POIs for each  training question.
We then create new training instances by randomly sampling $N$ non-answer POIs from the top-$k$ retrieved POIs, and use these to continue training the model.

\myparagraph{Inference}
Before inference, we disable the question encoder and generate representations of all POIs using the POI encoder only, and store and index them (as shown in the orange part in \figref{fig:model}).
During inference, the generated POI representations are loaded into memory.
Given a question $q$ at run-time, we encode it using the question encoder, score all candidates using the pre-computed representations, and return the top-$k$ results.

\section{Experimental Setup}

In this section, we introduce the dataset, baselines, and implementation details of our model.

\subsection{Dataset}
We use the TourismQA \cite{contractor2019large} dataset, which comprises over 47,000 real-world POI question--answer pairs from 50 cities across the globe. 
These questions are genuine queries submitted to a trip advisor website,\footnote{\url{https://www.tripadvisor.in}} and the answers are real-world responses that have been chosen and authenticated by annotators.
The average length of the questions is 87.48 tokens (separated by whitespace). And on average, there are 3.63 POIs as ground truth answers for each question. 
The dataset contains roughly 114,000 candidate POIs altogether, each with a collection of reviews and metadata such as geo-coordinates and type (restaurant, attraction, or hotel).

We follow \citet{contractor2019large} in dividing the dataset into a 9:1 train--test split, and constructing a search space by including POIs located in the same city as the ground truth POIs, resulting in an average of approximately 5,300 candidate POIs per question. 
We believe one reason for earlier work to build the candidate pool within a city was that their methods struggled with a large candidate pool. 
However, in real-world scenarios, the ground truth answer is concealed, and the candidate pool may be extensive, encompassing all POIs in the database. 
Therefore, we established a new evaluation setting in which the search space comprises all POIs in the world. 
We refer to this new setting as \emph{global} evaluation (114,000 candidates), and the previous one as \emph{local} evaluation (5,300 candidates).

\subsection{Evaluation Metrics}
Following \citet{contractor2019large}, we evaluate using Accuracy@$N\in\{3,5,30\}$ and mean reciprocal rank (MRR) for local evaluation, and use Accuracy@$N\in\{5,30,100\}$ for global evaluation.
For Accuracy@$N$, if the top-$N$ predictions have a non-empty intersection with the answer POI set, the results are considered to be correct.
For MRR, we return the reciprocal rank of the first positive answer POI per question, and average over the questions.

\begin{table*}[t]
    \small
	\centering
	\begin{tabular}{lccccccccc}
		\toprule
		\multirow{2}{*}{Model} & \multicolumn{4}{c}{Local} && \multicolumn{4}{c}{Global} \\
          \cmidrule{2-5}
          \cmidrule{7-10}
        & Acc@3 & Acc@5 & Acc@30 & MRR && Acc@5 & Acc@30 & Acc@100 & MRR\\
		\midrule
          SD & \z0.83 & \z1.11 & \z5.62 & 0.011    && \z0.83 & \z1.11 & \z5.62 & 0.011 \\
          BM25 & \z5.59 & \z8.29 & 16.33 & 0.061 &&  \z0.44 & \z1.86 & \z3.68 & 0.014 \\
          \midrule
          CRQA & 16.89 &23.75 &52.51 &0.159  &&-- &-- &--  &--   \\
          ST+CRQA & 19.37 &26.23 &\underline{56.33} &0.175 &&--  &--  &--  &--   \\
          CSRQA & 21.44 &28.20 &52.65 &0.186 &&--  &--  &--  &--   \\
          ST+CSRQA & \underline{22.41} &\underline{28.99} & 52.65 &\underline{0.193} &&--  &--  &--  &--   \\
          \midrule
        \lamb & \textbf{24.83} & \textbf{32.51} & \textbf{60.92} & \textbf{0.220} && \textbf{14.07} & \textbf{32.87} & 49.08 & \textbf{0.101} \\ 
        $-$Phase 2 & 22.49 & 29.35 & 59.20 & 0.201 && 13.22 & 31.89 & \textbf{49.68} & 0.094 \\
        $-$\selsum & 23.90 & 31.20 & 60.52 & 0.216 && 13.28 & 32.12 & 48.63 & 0.096 \\ 
        $-$$E_{loc}$ & 23.68 & 31.30 & 60.52 & 0.215 && \z9.59 & 24.52 & 40.03 & 0.071 \\
		\bottomrule
	\end{tabular}
	\caption{Overall evaluation on the TourismQA dataset. The second block of results are based on the TourismQA paper, wherein the best results are \underline{underlined}, and ``ST'' denotes the spatial--textual module. The overall best results are in \textbf{bold}. The third block presents the results for the full \lamb model, and also with module ablation.}
	\label{tab:result}
\end{table*}

\subsection{Baselines}

We compare ourselves against four baselines, as detailed below.

\myparagraph{Sort by Distance (SD)}: Given all tagged locations with geo-coordinates in the question, we rank POIs by the minimal distance from the tagged locations.

\myparagraph{BM25}: We represent each POI by its combined reviews, and index them using Apache Lucene. Then questions are used as a query to compute BM25 scores for all POIs.

\myparagraph{Cluster-Select-Rerank (``CSR'') Model} \cite{contractor2019large}, which consists of three components: 
(1) a clustering module that clusters reviews for each POI and selects representative reviews;
(2) a Duet \cite{Duet} retrieval model that selects the best 30 candidate POIs; and
(3) a QA-style re-ranker that scores and re-ranks the selected POIs.
Note that the cluster module is used to pre-process the POIs, and the selection and re-ranking modules are trained separately and pipelined.

\myparagraph{Spatial-Textual CSR} \cite{contractor2021joint}, which adds a self-attention based geospatial reasoner to the CSR model, and ranks POIs based on the weighted sum of scores from the geo-spatial reasoner and CSR.

\subsection{\lamb Implementation Details}
We implement our model in PyTorch, and use the HuggingFace \cite{wolf-etal-2020-transformers} implementation of DistilBERT \cite{DistilBERT} as the textual encoder.
The location module is comprised of two transformer blocks that are initialized using the first two blocks of a pre-trained DistilBERT model. 
We continued pre-training for 3 epochs using triplet loss to force the model to learn more spatial information, as described in \secref{sec:location}.   
During this process, we set the batch size to 8, learning rate to 2e-5, and the max sequence length to 64. 

For the main model of  \lamb, the maximum length (in subtokens) for both questions and reviews is set to 256.
For training, we use a linear learning rate scheduler with an initial learning rate of 2e-5, and the Adam optimizer with default hyperparameters.
For each training instance, we use a single positive POI and varying numbers of negatives.
We set the batch size to 8 and train for 10 epochs: 5 epochs of phase 1 (easy and medium negatives), and 5 epochs of phase 2 (medium and hard negatives).
All experiments were run on a single Nvidia A100 40GB GPU for about 8 hours.

\begin{table}[t]
    \small
	\centering
	\begin{tabular}{lrcrr}
          \toprule
          \multirow{2}{*}{Model} & Training && \multicolumn{2}{c}{Inference} \\
          \cmidrule{2-2}
          \cmidrule{4-5}
         & \multicolumn{1}{c}{Time (h)} && \multicolumn{1}{c}{\#Cand} & \multicolumn{1}{c}{Time (h)} \\
          \midrule
          CRQA ($\pm$ST)  & 360\phantom{+}  && 5.3k & 64  \\
          CSRQA ($\pm$ST) & 360+ && 30 & 2--3 \\
          \lamb  & 10\phantom{+} && 115k & 0.15 \\
          \bottomrule
	\end{tabular}
	\caption{Runtime comparison, based on a single Nvidia V100 GPU. ``\#Cand'' indicates the number of candidate POIs. For CSRQA, time was estimated by summing the times of the component models.}
	\label{tab:time}
\end{table}

\section{Results and Analysis}

\tabref{tab:result} shows the overall performance of the baselines and our proposed model.
We can see that the sparse-vector retrieval (BM25) and distance-based retrieval (SD) models in the first block of the table perform extremely poorly,  demonstrating the difficulty of the task.
In contrast, the textual-only pipelined models (CRQA and CSRQA) in the second block improve overall performance substantially, and adding the spatial reasoning sub-network (``ST+'') boosts results again.
Note that, since CSRQA is pipelined with a selection model that selects the top-30 results, the spatial-textual module cannot improve Accuracy@30 further. 

Compared to the baselines in blocks one and two, our model, \lamb, achieves the state-of-the-art across all metrics. 
To better understand the impact of different components of our model, we conducted an ablation study by separately removing the training phase 2, review selection and summarization modules, and location module. 
Overall, the performance dropped when one of these modules or strategies was removed, but still outperformed the previous state-of-the-art. 
Specifically, removing training phase 2 had a relatively large impact on local evaluation, which we attribute to the process of training to distinguish hard negatives. 
Removing the location module greatly impacted the global evaluation, demonstrating the effectiveness of the location module, particularly when candidates are from around the globe.

Based on our analysis, there are three main reasons why \lamb outperforms previous models: 
(1) training and inference are end-to-end, avoiding error propagation due to pipelining, as with CSRQA;
(2) our use of pre-trained language models as the textual encoder, outperforming static word embeddings or training encoders from scratch; and
(3) learning location encodings separately and fusing them with textual representations, providing a soft distance computing method.
We provide a comparison between our location module design and other straightforward geo-coordinate-based location/distance modules in \appref{sec:distance}.
From this, we can conclude that compared to strategies that encode geo-coordinates directly, a pretrained location name module better captures spatial information.

\begin{table}[t]
	\small
	\centering
	\begin{tabular}{llcccc}
		\toprule
		Loc Module & Acc@5 & Acc@30 & Acc@100 & MRR \\
		\midrule
        w/o Loc  &  \z9.59 & 24.52 & 40.03 & 0.071 \\
        2-$l$ PLM  &  \z9.91 & 25.47 & 41.64 & 0.075 \\
        1-$l$ PLM-Loc &  12.86 & 30.67 & 46.28 & 0.091 \\ 
        2-$l$ PLM-Loc &  \textbf{14.07} & \textbf{32.87} & \textbf{49.08} & \textbf{0.101} \\ 
        4-$l$ PLM-Loc &  10.92 & 27.78 & 43.24 & 0.081 \\ 
		\bottomrule
	\end{tabular}
	\caption{Results with  different location module settings. ``PLM'' = use PLM directly; ``PLM-Loc'' = continue to pretrain PLM on location names; and ``$N$-$l$'' = use $N$ transformer blocks.}
	\label{tab:loc}
\end{table}

\subsection{Efficiency Comparison}
We analyze the computational requirements of the models in \tabref{tab:time}.
\lamb is more time efficient than the previously-proposed neural models, requiring around 5\% of the training time, and $<$10\% of the inference time.
It is also able to handle a much larger candidate pool (in the millions of candidates) compared to C($\pm$S)RQA (in the tens or thousands of candidates).
Further analysis of efficiency and usability is provided in \appref{sec:usability}.

\begin{table*}[t]
    \small
	\centering
	\begin{tabular}{lccccccccc}
		\toprule
		\multirow{2}{*}{\#HN} & \multicolumn{4}{c}{Local} && \multicolumn{4}{c}{Global} \\
          \cmidrule{2-5}
          \cmidrule{7-10}
        & Acc@3 & Acc@5 & Acc@30 & MRR && Acc@5 & Acc@30 & Acc@100 & MRR\\
		\midrule
        0 & 16.41 & 22.59 & 51.05 & 0.159 && 19.51 & \textbf{44.70} & \textbf{64.91} & \textbf{0.137} \\    
        1 & 15.51 & 21.34 & 51.67 & 0.154 && 16.97 & 43.06 & 62.19 & 0.123 \\    
        4 & 20.55 & 27.29 & 52.41 & 0.188 && \textbf{20.13} & 40.43 & 56.27 & 0.142 \\    
        8 & 23.99 & 30.58 & 59.47 & 0.213 && 17.37 & 37.78 & 54.20 & 0.124\\    
        12 & \textbf{24.83} & \textbf{32.51} & \textbf{60.92} & \textbf{0.220} && 14.07 & 32.87 & 49.08 & 0.101 \\ 
        15 & 24.06 & 31.44 & 60.70 & 0.221 && \z8.10 & 21.34 & 36.18 & 0.063 \\    
		\bottomrule
	\end{tabular}
	\caption{Results with differing numbers of easy/hard negatives, total negatives = 15. \#HN: number of hard negatives.}
	\label{tab:neg}
\end{table*}
 \begin{table*}[t]
    \small
	\centering
	\begin{tabular}{lccccccccc}
		\toprule
		\multirow{2}{*}{\#Phase 1,2} & \multicolumn{4}{c}{Local} && \multicolumn{4}{c}{Global} \\
          \cmidrule{2-5}
          \cmidrule{7-10}
        & Acc@3 & Acc@5 & Acc@30 & MRR && Acc@5 & Acc@30 & Acc@100 & MRR\\
		\midrule
        10, 0 & 22.49 & 29.35 & 59.20 & 0.201 && 13.22 & 31.89 & 49.68 & 0.094 \\
        8, 2 & 23.08 & 30.22 & 60.33 & 0.209 && 13.48 & 32.48 & \textbf{50.65} & 0.096 \\
        5, 5 & \textbf{24.83} & \textbf{32.51} & \textbf{60.92} & \textbf{0.220} && \textbf{14.07} & \textbf{32.87} & 49.08 & \textbf{0.101} \\ 
        2, 8 & 24.49 & 31.88 & 60.08 & 0.219 && 13.05 & 31.18 & 46.04 & 0.095 \\
        0, 10 & 21.73 & 28.28 & 54.15 & 0.198 && 11.70 & 28.74 & 42.82 & 0.085 \\
		\bottomrule
	\end{tabular}
	\caption{Results with varied epochs in two-phase training, using 10 total training epochs.}
	\label{tab:two_phase}
\end{table*}

\subsection{Ablation Study on Model Training}
To further understand how different model training options affect the results, we conduct several additional experiments and discuss our findings below.

\paragraph{Location Module Analysis}\label{sec:loc}
In this section, we compare various settings of location modules as shown in Table \ref{tab:loc}. 
The table indicates that continuous pretraining of a PLM on location names significantly enhances the module's ability to capture geo-location and distance. 
Furthermore, using two transformer blocks is sufficient to encode multi-granularity location names, whereas more or fewer layers may lead to overfitting or underfitting.

\paragraph{Effectiveness of Negative Examples}\label{sec:neg}
To investigate the effectiveness of the type and number of negative examples during training, we kept the total number of negatives constant at 15 while varying the mix of easy and hard negatives (as presented in \tabref{tab:neg}).
As we increase the number of hard negatives, the global evaluation results deteriorate while the local evaluation results improve. 
This implies that training with easy negatives is more appropriate when the target city or area is unconstrained. 
The best local evaluation results were achieved when using 12/15 hard negatives, indicating that easy negatives are still necessary for learning general location constraints.
We further investigated varying the total number of negatives for contrastive learning, as presented in \tabref{tab:more_neg} in the Appendix. 
Our findings indicate that the more negatives we have in each training instance, the better the model performs, but that the relative improvement plateaus beyond around 30.

\paragraph{Two-Phase Training Strategy}
We conducted experiments with different epoch configurations for our two-phase training strategy, as detailed in \tabref{tab:two_phase}. 
Our results indicate that both phase 1 and phase 2 are essential, aligning with the assumptions stated in \secref{sec:training}. 
Furthermore, we found that commencing phase 2 training at the midway point was particularly effective.

\subsection{Human Evaluation}\label{sec:human}
To further investigate the dataset and have a better sense of the overall performance of \lamb, we conducted a small-scale human evaluation.
We randomly choose 100 questions from the test set and manually evaluate the top-3 predictions for relevance based on \lamb as presented in \tabref{tab:result}.
For this small question set, our estimate of the true Accuracy@3 is around 75\%, as compared to the automatic evaluation result of 24\%.
This is consistent with the human evaluation results reported in \cite{contractor2021joint}, and points to the issue of low label-recall in the dataset: while a given POI may not have been selected by the user who issued the original question, it may well have satisfied the constraints described in the question.

\subsection{How ChatGPT Performs on TourismQA}\label{sec:chatgpt}
During the writing of this paper, ChatGPT (i.e.\ GPT3.5) was released. 
We manually tested 100 questions from \secref{sec:human} by inputting them directly into ChatGPT (GPT-3.5-turbo on 20-March-2023) and getting a single response.\footnote{Questions and responses are released together with the source code.} 
The results show that out of the 100 questions, 91 received recommendations for points of interest or areas. 
However, only 14 of those replies match the ground truth answers, which is lower than our model's performance of 24. 
We believe that the main reason for this discrepancy is due to differences in the POI databases. 
The replies from ChatGPT were well-organized and logical, and could even answer many details in the questions beyond the capabilities of our model.

However, we observed that ChatGPT failed to provide an output in many cases:
among the 100 replies, sentences such as \ex{As an AI language model, I don't have personal experience in ... } appeared 36 times, while other outputs like \ex{I can recommend that you check out the reviews on websites like TripAdvisor or Booking.com} appeared 13 times. 
Additionally, ChatGPT tended to recommend popular places, with the word \ex{popular} appearing 44 times in replies, despite not being mentioned in any of the questions. 
We observed further bias in ChatGPT's recommendations. 
For example, it recommended \ex{Shake Shack} nine times in response to fast food requests, but never mentioned other international fast-food chains or local chains, 
even when questions specifically asked for fast food with regional characteristics.

Lastly, ChatGPT's database is not up-to-date, as also mentioned in its replies. 
Since OpenAI did not provide full training details, the cost of updating the database, including fine-tuning the model, is unclear. 
In summary, there is still a real need for a comprehensive recommendation system that can be combined with up-to-date website information.

\section{Related Work}

\paragraph{Geo-Spatial Question Anwering}
There has been a strong focus in the literature on component geospatial tasks such as geo-parsing (toponym recognition and disambiguation) \cite{karimzadeh2019geotxt,neuraltpr}, geo-tagging (tagging toponyms with geographic metadata) \cite{compton2014geotagging,middleton2018location}, geospatial information retrieval \cite{purves2018geographic}, and geospatial question analysis \cite{DBLP:conf/agile/HamzeiLVB0T19}.

Based on the type of question, existing work on geospatial QA (``GeoQA'')
can be classified into four types \cite{mai2021geographic}: 
(1) factoid GQA \cite{li2021neural,DBLP:conf/www/HamzeiT022}, focusing on answering questions with geographic factoids; 
(2) geo-analytical QA \cite{scheider2020geo,agile-giss-1-23-2020}, focusing on questions with complex spatial analytical intent; 
(3) visual GQA \cite{lobry2020rsvqa,janowicz2020geoai}, linking questions to an image or video;
and (4) scenario-based GQA \cite{DBLP:conf/emnlp/HuangSLWCZDQ19,contractor2019large}, which associates questions with a scenario described with a map or paragraph of text. 
Our work corresponds to the last type, and unlike most other work, we do not rely on task-specific query languages or annotations, and focus more on NLP and IR modeling.

\paragraph{Point-of-Interest (POI) Recommendation}
POI recommendation systems have a wide range of applications such as online navigation applications \cite{PALM,IncreSTGL},
personalized recommendation systems in location-based social networks \cite{PRME,Next-POI},
and trip or accommodation advisory systems \cite{GPQ,contractor2019large}.
In this research, we focus on POI recommendation incorporating both structured information (such as geo-coordinates) and unstructured information (such as textual descriptions).
Previous work has explored efficient spatial indexing based on specialized data structures, with textual information as sparse vectors or filters \cite{preference-query0,GPQ}.
Recent work \cite{contractor2019large, contractor2021joint} has focused on latent textual representations, which is highly relevant here.

\paragraph{Textual Encoding and Document Retrieval}
Pretrained language models (PLMs) have led to great successes across many NLP tasks \cite{BERT,Roberta,XLNet,BART,DeBERTa,ELECTRA}. 
In the field of QA, PLMs have been used to generate representations of questions and documents \cite{BERT-indexer,SG-Net}.
In this work, we use DistilBERT \cite{DistilBERT} as our textual encoder, as it is more efficient than BERT and retains much of its expressivity. 

Document retrieval has become a mainstay of research in IR and QA.
Recently, IR has increasingly moved towards dense vector retrieval methods \cite{DBLP:conf/iclr/DasDZM19,DBLP:conf/acl/SeoLKPFH19,DBLP:conf/eacl/XiongWW21}.
In particular, \citet{DPR} proposed DPR based on a dual-encoder approach, and attained impressive results on multiple open-domain question answering benchmarks. 
Inspired by this, we adopt a bi-encoder framework.

\section{Conclusion}
We have proposed the \lamb model, a location-aware bi-encoder model for answering POI recommendation questions.
Experiments on a recently-released tourism question-answering dataset show that our model surpasses existing spatial-textual reasoning models across all metrics.
Experiments over \lamb's components and based on changing up the training strategy show the effectiveness of the different design choices used in \lamb.
Finally, we analyzed the training and inference efficiency, and demonstrated that our model is resource-efficient at training and inference time, suggesting it can be deployed in real-world tourism applications.

\section*{Limitations}
Although we have achieved results that significantly outperform the current state-of-the-art, 
our work still has some limitations.
First, as demonstrated in \secref{sec:human} and in the earlier work of \citet{contractor2021joint}, the TourismQA dataset was
collected semi-automatically, and the gold labels have high precision
but low recall. Hence any results on this dataset are likely an
underestimate of the true model performance.
While we currently use the Haversine formula to compute the distance between two locations and supervise the pre-training of the location module, we recognize that this calculation may not reflect the actual distance between two places, taking into account the route direction and vertical height difference. 
In light of the city's urban design, the Manhattan distance might better represent the true distance between two locations within a city. 
Additionally, POI density could be a factor that influences user choice
in real life, in that people may be more inclined to go to locations
with a higher density of restaurants to eat (in order to have more
options if a given restaurant doesn't live up to their expectations),
rather than travel far to a remote place without other options in the
local vicinity. For hotels, on the other hand, some users may prefer
privacy and a lower density.
Such extra-linguistic features are not explicitly captured in our model.

\bibliography{custom}

\begin{thebibliography}{52}
\expandafter\ifx\csname natexlab\endcsname\relax\def\natexlab#1{#1}\fi

\bibitem[{Bražinskas et~al.(2021)Bražinskas, Lapata, and Titov}]{SelSum}
Arthur Bražinskas, Mirella Lapata, and Ivan Titov. 2021.
\newblock Learning opinion summarizers by selecting informative reviews.
\newblock In \emph{Proceedings of the Conference on Empirical Methods in
  Natural Language Processing (EMNLP)}.

\bibitem[{Clark et~al.(2020)Clark, Luong, Le, and Manning}]{ELECTRA}
Kevin Clark, Minh{-}Thang Luong, Quoc~V. Le, and Christopher~D. Manning. 2020.
\newblock \href {https://openreview.net/forum?id=r1xMH1BtvB} {{ELECTRA:}
  pre-training text encoders as discriminators rather than generators}.
\newblock In \emph{Proceedings of the 8th International Conference on Learning
  Representations}. OpenReview.net.

\bibitem[{Compton et~al.(2014)Compton, Jurgens, and
  Allen}]{compton2014geotagging}
Ryan Compton, David Jurgens, and David Allen. 2014.
\newblock \href {https://doi.org/10.1109/BigData.2014.7004256} {Geotagging one
  hundred million twitter accounts with total variation minimization}.
\newblock In \emph{Proceedings of the 2014 {IEEE} International Conference on
  Big Data}, pages 393--401. {IEEE} Computer Society.

\bibitem[{Contractor et~al.(2021{\natexlab{a}})Contractor, Goel, and
  Singla}]{contractor2021joint}
Danish Contractor, Shashank Goel, and Parag Singla. 2021{\natexlab{a}}.
\newblock Joint spatio-textual reasoning for answering tourism questions.
\newblock In \emph{Proceedings of the Web Conference 2021}, pages 1978--1989.

\bibitem[{Contractor et~al.(2021{\natexlab{b}})Contractor, Shah, Partap,
  Singla, and Mausam}]{contractor2019large}
Danish Contractor, Krunal Shah, Aditi Partap, Parag Singla, and Mausam.
  2021{\natexlab{b}}.
\newblock \href {https://doi.org/10.1145/3459637.3482320} {Answering
  poi-recommendation questions using tourism reviews}.
\newblock In \emph{Proceedings of the 30th {ACM} International Conference on
  Information and Knowledge Management}, pages 281--291. {ACM}.

\bibitem[{Cui et~al.(2020)Cui, Xu, Fei, Cai, Cao, Zhang, and
  Chen}]{cui2020personalized}
Zhihua Cui, Xianghua Xu, Xue Fei, Xingjuan Cai, Yang Cao, Wensheng Zhang, and
  Jinjun Chen. 2020.
\newblock Personalized recommendation system based on collaborative filtering
  for {IoT} scenarios.
\newblock \emph{IEEE Transactions on Services Computing}, 13(4):685--695.

\bibitem[{Das et~al.(2019)Das, Dhuliawala, Zaheer, and
  McCallum}]{DBLP:conf/iclr/DasDZM19}
Rajarshi Das, Shehzaad Dhuliawala, Manzil Zaheer, and Andrew McCallum. 2019.
\newblock \href {https://openreview.net/forum?id=HkfPSh05K7} {Multi-step
  retriever-reader interaction for scalable open-domain question answering}.
\newblock In \emph{Proceedings of the 7th International Conference on Learning
  Representations}. OpenReview.net.

\bibitem[{de~Almeida and Rocha{-}Junior(2015)}]{preference-query0}
Jo{\~{a}}o Paulo~Dias de~Almeida and Jo{\~{a}}o~B. Rocha{-}Junior. 2015.
\newblock \href
  {https://sol.sbc.org.br/journals/index.php/jidm/article/view/1568} {Top-k
  spatial keyword preference query}.
\newblock \emph{Journal of Information and Data Management}, 6(3):162--177.

\bibitem[{Devlin et~al.(2019)Devlin, Chang, Lee, and Toutanova}]{BERT}
Jacob Devlin, Ming-Wei Chang, Kenton Lee, and Kristina Toutanova. 2019.
\newblock \href {https://doi.org/10.18653/v1/N19-1423} {{BERT}: Pre-training of
  deep bidirectional transformers for language understanding}.
\newblock In \emph{Proceedings of the 2019 Conference of the North {A}merican
  Chapter of the Association for Computational Linguistics: Human Language
  Technologies, Volume 1 (Long and Short Papers)}, pages 4171--4186,
  Minneapolis, Minnesota. Association for Computational Linguistics.

\bibitem[{Feng et~al.(2015)Feng, Li, Zeng, Cong, Chee, and Yuan}]{PRME}
Shanshan Feng, Xutao Li, Yifeng Zeng, Gao Cong, Yeow~Meng Chee, and Quan Yuan.
  2015.
\newblock \href {http://ijcai.org/Abstract/15/293} {Personalized ranking metric
  embedding for next new {POI} recommendation}.
\newblock In \emph{Proceedings of the Twenty-Fourth International Joint
  Conference on Artificial Intelligence}, pages 2069--2075. {AAAI} Press.

\bibitem[{Hamid et~al.(2021)Hamid, Albahri, Alwan, Al-Qaysi, Albahri, Zaidan,
  Alnoor, Alamoodi, and Zaidan}]{hamid2021smart}
Rula~A Hamid, Ahmed~Shihab Albahri, Jwan~K Alwan, ZT~Al-Qaysi, Osamah~Shihab
  Albahri, AA~Zaidan, Alhamzah Alnoor, AH~Alamoodi, and BB~Zaidan. 2021.
\newblock How smart is e-tourism? a systematic review of smart tourism
  recommendation system applying data management.
\newblock \emph{Computer Science Review}, 39:100337.

\bibitem[{Hamzei et~al.(2019)Hamzei, Li, Vasardani, Baldwin, Winter, and
  Tomko}]{DBLP:conf/agile/HamzeiLVB0T19}
Ehsan Hamzei, Haonan Li, Maria Vasardani, Timothy Baldwin, Stephan Winter, and
  Martin Tomko. 2019.
\newblock \href {https://doi.org/10.1007/978-3-030-14745-7\_1} {Place questions
  and human-generated answers: {A} data analysis approach}.
\newblock In \emph{Proceedings of the 22nd {AGILE} Conference on Geographic
  Information Science}, pages 3--19. Springer.

\bibitem[{Hamzei et~al.(2022)Hamzei, Tomko, and
  Winter}]{DBLP:conf/www/HamzeiT022}
Ehsan Hamzei, Martin Tomko, and Stephan Winter. 2022.
\newblock \href {https://doi.org/10.1145/3485447.3511933} {Translating
  place-related questions to geosparql queries}.
\newblock In \emph{Proceedings of the {ACM} Web Conference 2022}, pages
  902--911. {ACM}.

\bibitem[{He et~al.(2021)He, Liu, Gao, and Chen}]{DeBERTa}
Pengcheng He, Xiaodong Liu, Jianfeng Gao, and Weizhu Chen. 2021.
\newblock \href {https://openreview.net/forum?id=XPZIaotutsD} {{DeBERTa}:
  decoding-enhanced {Bert} with disentangled attention}.
\newblock In \emph{Proceedings of the 9th International Conference on Learning
  Representations}. OpenReview.net.

\bibitem[{Hofmann et~al.(2022)Hofmann, Glava{\v{s}}, Ljube{\v{s}}i{\'c},
  Pierrehumbert, and Sch{\"u}tze}]{hofmann2022geographic}
Valentin Hofmann, Goran Glava{\v{s}}, Nikola Ljube{\v{s}}i{\'c}, Janet~B
  Pierrehumbert, and Hinrich Sch{\"u}tze. 2022.
\newblock Geographic adaptation of pretrained language models.
\newblock \emph{arXiv preprint arXiv:2203.08565}.

\bibitem[{Huang et~al.(2019)Huang, Shen, Li, Wei, Cheng, Zhou, Dai, and
  Qu}]{DBLP:conf/emnlp/HuangSLWCZDQ19}
Zixian Huang, Yulin Shen, Xiao Li, Yu{'}ang Wei, Gong Cheng, Lin Zhou, Xinyu
  Dai, and Yuzhong Qu. 2019.
\newblock \href {https://doi.org/10.18653/v1/D19-1597} {{G}eo{SQA}: A benchmark
  for scenario-based question answering in the geography domain at high school
  level}.
\newblock In \emph{Proceedings of the 2019 Conference on Empirical Methods in
  Natural Language Processing and the 9th International Joint Conference on
  Natural Language Processing (EMNLP-IJCNLP)}, pages 5866--5871, Hong Kong,
  China. Association for Computational Linguistics.

\bibitem[{Janowicz et~al.(2020)Janowicz, Gao, McKenzie, Hu, and
  Bhaduri}]{janowicz2020geoai}
Krzysztof Janowicz, Song Gao, Grant McKenzie, Yingjie Hu, and Budhendra
  Bhaduri. 2020.
\newblock \href {https://doi.org/10.1080/13658816.2019.1684500} {{GeoAI}:
  Spatially explicit artificial intelligence techniques for geographic
  knowledge discovery and beyond}.
\newblock \emph{International Journal of Geographic Information Science}, pages
  625--636.

\bibitem[{Jiao et~al.(2019)Jiao, Yin, Shang, Jiang, Chen, Li, Wang, and
  Liu}]{Tinybert}
Xiaoqi Jiao, Yichun Yin, Lifeng Shang, Xin Jiang, Xiao Chen, Linlin Li, Fang
  Wang, and Qun Liu. 2019.
\newblock Tinybert: Distilling bert for natural language understanding.
\newblock \emph{arXiv preprint arXiv:1909.10351}.

\bibitem[{Johnson et~al.(2021)Johnson, Douze, and J{\'{e}}gou}]{FAISS}
Jeff Johnson, Matthijs Douze, and Herv{\'{e}} J{\'{e}}gou. 2021.
\newblock \href {https://doi.org/10.1109/TBDATA.2019.2921572} {Billion-scale
  similarity search with {GPUs}}.
\newblock \emph{{IEEE} Trans. Big Data}, 7(3):535--547.

\bibitem[{Karimzadeh et~al.(2019)Karimzadeh, Pezanowski, MacEachren, and
  Wallgr{\"u}n}]{karimzadeh2019geotxt}
Morteza Karimzadeh, Scott Pezanowski, Alan~M MacEachren, and Jan~O
  Wallgr{\"u}n. 2019.
\newblock Geotxt: A scalable geoparsing system for unstructured text
  geolocation.
\newblock \emph{Transactions in GIS}, 23(1):118--136.

\bibitem[{Karpukhin et~al.(2020)Karpukhin, Oguz, Min, Lewis, Wu, Edunov, Chen,
  and Yih}]{DPR}
Vladimir Karpukhin, Barlas Oguz, Sewon Min, Patrick Lewis, Ledell Wu, Sergey
  Edunov, Danqi Chen, and Wen-tau Yih. 2020.
\newblock \href {https://doi.org/10.18653/v1/2020.emnlp-main.550} {Dense
  passage retrieval for open-domain question answering}.
\newblock In \emph{Proceedings of the 2020 Conference on Empirical Methods in
  Natural Language Processing (EMNLP)}, pages 6769--6781, Online. Association
  for Computational Linguistics.

\bibitem[{Kwiatkowski et~al.(2019)Kwiatkowski, Palomaki, Redfield, Collins,
  Parikh, Alberti, Epstein, Polosukhin, Devlin, Lee, Toutanova, Jones, Kelcey,
  Chang, Dai, Uszkoreit, Le, and Petrov}]{NQ}
Tom Kwiatkowski, Jennimaria Palomaki, Olivia Redfield, Michael Collins, Ankur
  Parikh, Chris Alberti, Danielle Epstein, Illia Polosukhin, Jacob Devlin,
  Kenton Lee, Kristina Toutanova, Llion Jones, Matthew Kelcey, Ming-Wei Chang,
  Andrew~M. Dai, Jakob Uszkoreit, Quoc Le, and Slav Petrov. 2019.
\newblock \href {https://doi.org/10.1162/tacl_a_00276} {Natural questions: A
  benchmark for question answering research}.
\newblock \emph{Transactions of the Association for Computational Linguistics},
  7:452--466.

\bibitem[{Lee et~al.(2019)Lee, Chang, and Toutanova}]{lee2019latent}
Kenton Lee, Ming-Wei Chang, and Kristina Toutanova. 2019.
\newblock \href {https://doi.org/10.18653/v1/P19-1612} {Latent retrieval for
  weakly supervised open domain question answering}.
\newblock In \emph{Proceedings of the 57th Annual Meeting of the Association
  for Computational Linguistics}, pages 6086--6096, Florence, Italy.
  Association for Computational Linguistics.

\bibitem[{Lewis et~al.(2020)Lewis, Liu, Goyal, Ghazvininejad, Mohamed, Levy,
  Stoyanov, and Zettlemoyer}]{BART}
Mike Lewis, Yinhan Liu, Naman Goyal, Marjan Ghazvininejad, Abdelrahman Mohamed,
  Omer Levy, Veselin Stoyanov, and Luke Zettlemoyer. 2020.
\newblock \href {https://doi.org/10.18653/v1/2020.acl-main.703} {{BART}:
  Denoising sequence-to-sequence pre-training for natural language generation,
  translation, and comprehension}.
\newblock In \emph{Proceedings of the 58th Annual Meeting of the Association
  for Computational Linguistics}, pages 7871--7880, Online. Association for
  Computational Linguistics.

\bibitem[{Li et~al.(2021)Li, Hamzei, Majic, Hua, Renz, Tomko, Vasardani,
  Winter, and Baldwin}]{li2021neural}
Haonan Li, Ehsan Hamzei, Ivan Majic, Hua Hua, Jochen Renz, Martin Tomko, Maria
  Vasardani, Stephan Winter, and Timothy Baldwin. 2021.
\newblock Neural factoid geospatial question answering.
\newblock \emph{Journal of Spatial Information Science}, 23:65--90.

\bibitem[{Li et~al.(2016)Li, Chen, Cong, Gu, and Yu}]{GPQ}
Miao Li, Lisi Chen, Gao Cong, Yu~Gu, and Ge~Yu. 2016.
\newblock \href {https://doi.org/10.1145/2983323.2983757} {Efficient processing
  of location-aware group preference queries}.
\newblock In \emph{Proceedings of the 25th {ACM} International Conference on
  Information and Knowledge Management}, pages 559--568. {ACM}.

\bibitem[{Liu et~al.(2019)Liu, Ott, Goyal, Du, Joshi, Chen, Levy, Lewis,
  Zettlemoyer, and Stoyanov}]{Roberta}
Yinhan Liu, Myle Ott, Naman Goyal, Jingfei Du, Mandar Joshi, Danqi Chen, Omer
  Levy, Mike Lewis, Luke Zettlemoyer, and Veselin Stoyanov. 2019.
\newblock \href {https://arxiv.org/abs/1907.11692} {{RoBERTa}: {A} robustly
  optimized {BERT} pretraining approach}.
\newblock \emph{ArXiv preprint}, abs/1907.11692.

\bibitem[{Lobry et~al.(2020)Lobry, Marcos, Murray, and Tuia}]{lobry2020rsvqa}
Sylvain Lobry, Diego Marcos, Jesse Murray, and Devis Tuia. 2020.
\newblock \href {https://doi.org/10.1109/TGRS.2020.2988782} {{RSVQA}: Visual
  question answering for remote sensing data}.
\newblock \emph{IEEE Transactions on Geoscience and Remote Sensing},
  58(12):8555--8566.

\bibitem[{Mai et~al.(2021)Mai, Janowicz, Zhu, Cai, and Lao}]{mai2021geographic}
Gengchen Mai, Krzysztof Janowicz, Rui Zhu, Ling Cai, and Ni~Lao. 2021.
\newblock Geographic question answering: Challenges, uniqueness,
  classification, and future directions.
\newblock \emph{AGILE: GIScience Series}, 2:1--21.

\bibitem[{Middleton et~al.(2018)Middleton, Kordopatis-Zilos, Papadopoulos, and
  Kompatsiaris}]{middleton2018location}
Stuart~E Middleton, Giorgos Kordopatis-Zilos, Symeon Papadopoulos, and Yiannis
  Kompatsiaris. 2018.
\newblock Location extraction from social media: Geoparsing, location
  disambiguation, and geotagging.
\newblock \emph{ACM Transactions on Information Systems (TOIS)}, 36(4):1--27.

\bibitem[{Mitra and Craswell(2019)}]{Duet}
Bhaskar Mitra and Nick Craswell. 2019.
\newblock \href {https://arxiv.org/abs/1903.07666} {An updated duet model for
  passage re-ranking}.
\newblock \emph{ArXiv preprint}, abs/1903.07666.

\bibitem[{Nogueira et~al.(2019)Nogueira, Yang, Lin, and Cho}]{BERT-indexer}
Rodrigo Nogueira, Wei Yang, Jimmy Lin, and Kyunghyun Cho. 2019.
\newblock \href {https://arxiv.org/abs/1904.08375} {Document expansion by query
  prediction}.
\newblock \emph{ArXiv preprint}, abs/1904.08375.

\bibitem[{Punjani et~al.(2018)Punjani, Singh, Both, Koubarakis, Angelidis,
  Bereta, Beris, Bilidas, Ioannidis, Karalis, and Lange}]{punjani2018template}
Dharmen Punjani, K~Singh, Andreas Both, Manolis Koubarakis, Ioannis Angelidis,
  Konstantina Bereta, Themis Beris, Dimitris Bilidas, T~Ioannidis, Nikolaos
  Karalis, and C.~Lange. 2018.
\newblock \href {https://doi.org/10.1145/3281354.3281362} {Template-based
  question answering over linked geospatial data}.
\newblock In \emph{Proceedings of the 12th Workshop on Geographic Information
  Retrieval}, page~7.

\bibitem[{Purves et~al.(2018)Purves, Clough, Jones, Hall, and
  Murdock}]{purves2018geographic}
Ross~S Purves, Paul Clough, Christopher~B Jones, Mark~H Hall, and Vanessa
  Murdock. 2018.
\newblock Geographic information retrieval: Progress and challenges in spatial
  search of text.
\newblock \emph{Foundations and Trends in Information Retrieval},
  12(2-3):164--318.

\bibitem[{Rajpurkar et~al.(2016)Rajpurkar, Zhang, Lopyrev, and Liang}]{SQuAD}
Pranav Rajpurkar, Jian Zhang, Konstantin Lopyrev, and Percy Liang. 2016.
\newblock \href {https://doi.org/10.18653/v1/D16-1264} {{SQ}u{AD}: 100,000+
  questions for machine comprehension of text}.
\newblock In \emph{Proceedings of the 2016 Conference on Empirical Methods in
  Natural Language Processing}, pages 2383--2392, Austin, Texas. Association
  for Computational Linguistics.

\bibitem[{Sanh et~al.(2019)Sanh, Debut, Chaumond, and Wolf}]{DistilBERT}
Victor Sanh, Lysandre Debut, Julien Chaumond, and Thomas Wolf. 2019.
\newblock \href {https://arxiv.org/abs/1910.01108} {Distilbert, a distilled
  version of {BERT:} smaller, faster, cheaper and lighter}.
\newblock \emph{ArXiv preprint}, abs/1910.01108.

\bibitem[{Scheider et~al.(2020)Scheider, Nyamsuren, Kruiger, and
  Xu}]{scheider2020geo}
Simon Scheider, Enkhbold Nyamsuren, Han Kruiger, and Haiqi Xu. 2020.
\newblock \href {https://doi.org/10.1080/17538947.2020.1738568} {Geo-analytical
  question-answering with {GIS}}.
\newblock \emph{International Journal of Digital Earth}, pages 1--14.

\bibitem[{Scherrer and Ljube{\v{s}}i{\'c}(2021)}]{scherrer2021social}
Yves Scherrer and Nikola Ljube{\v{s}}i{\'c}. 2021.
\newblock Social media variety geolocation with {GeoBERT}.
\newblock In \emph{Proceedings of the Eighth Workshop on NLP for Similar
  Languages, Varieties and Dialects}. The Association for Computational
  Linguistics.

\bibitem[{Seo et~al.(2017)Seo, Kembhavi, Farhadi, and Hajishirzi}]{bidaf}
Min~Joon Seo, Aniruddha Kembhavi, Ali Farhadi, and Hannaneh Hajishirzi. 2017.
\newblock \href {https://openreview.net/forum?id=HJ0UKP9ge} {Bidirectional
  attention flow for machine comprehension}.
\newblock In \emph{Proceedings of the 5th International Conference on Learning
  Representations}. OpenReview.net.

\bibitem[{Seo et~al.(2019)Seo, Lee, Kwiatkowski, Parikh, Farhadi, and
  Hajishirzi}]{DBLP:conf/acl/SeoLKPFH19}
Minjoon Seo, Jinhyuk Lee, Tom Kwiatkowski, Ankur Parikh, Ali Farhadi, and
  Hannaneh Hajishirzi. 2019.
\newblock \href {https://doi.org/10.18653/v1/P19-1436} {Real-time open-domain
  question answering with dense-sparse phrase index}.
\newblock In \emph{Proceedings of the 57th Annual Meeting of the Association
  for Computational Linguistics}, pages 4430--4441, Florence, Italy.
  Association for Computational Linguistics.

\bibitem[{Sun et~al.(2020)Sun, Yu, Song, Liu, Yang, and Zhou}]{Mobilebert}
Zhiqing Sun, Hongkun Yu, Xiaodan Song, Renjie Liu, Yiming Yang, and Denny Zhou.
  2020.
\newblock {MobileBERT}: a compact task-agnostic bert for resource-limited
  devices.
\newblock \emph{arXiv preprint arXiv:2004.02984}.

\bibitem[{Wang et~al.(2020{\natexlab{a}})Wang, Hu, and Joseph}]{neuraltpr}
Jimin Wang, Yingjie Hu, and Kenneth Joseph. 2020{\natexlab{a}}.
\newblock {NeuroTPR}: A neuro-net toponym recognition model for extracting
  locations from social media messages.
\newblock \emph{Transactions in GIS}, 24(3):719--735.

\bibitem[{Wang et~al.(2020{\natexlab{b}})Wang, Wei, Dong, Bao, Yang, and
  Zhou}]{Minilm}
Wenhui Wang, Furu Wei, Li~Dong, Hangbo Bao, Nan Yang, and Ming Zhou.
  2020{\natexlab{b}}.
\newblock {MiniLM}: Deep self-attention distillation for task-agnostic
  compression of pre-trained transformers.
\newblock \emph{Advances in Neural Information Processing Systems},
  33:5776--5788.

\bibitem[{Wolf et~al.(2020)Wolf, Debut, Sanh, Chaumond, Delangue, Moi, Cistac,
  Rault, Louf, Funtowicz, Davison, Shleifer, von Platen, Ma, Jernite, Plu, Xu,
  Le~Scao, Gugger, Drame, Lhoest, and Rush}]{wolf-etal-2020-transformers}
Thomas Wolf, Lysandre Debut, Victor Sanh, Julien Chaumond, Clement Delangue,
  Anthony Moi, Pierric Cistac, Tim Rault, Remi Louf, Morgan Funtowicz, Joe
  Davison, Sam Shleifer, Patrick von Platen, Clara Ma, Yacine Jernite, Julien
  Plu, Canwen Xu, Teven Le~Scao, Sylvain Gugger, Mariama Drame, Quentin Lhoest,
  and Alexander Rush. 2020.
\newblock \href {https://doi.org/10.18653/v1/2020.emnlp-demos.6} {Transformers:
  State-of-the-art natural language processing}.
\newblock In \emph{Proceedings of the 2020 Conference on Empirical Methods in
  Natural Language Processing: System Demonstrations}, pages 38--45, Online.
  Association for Computational Linguistics.

\bibitem[{Xiong et~al.(2021{\natexlab{a}})Xiong, Xiong, Li, Tang, Liu, Bennett,
  Ahmed, and Overwijk}]{DBLP:conf/iclr/XiongXLTLBAO21}
Lee Xiong, Chenyan Xiong, Ye~Li, Kwok{-}Fung Tang, Jialin Liu, Paul~N. Bennett,
  Junaid Ahmed, and Arnold Overwijk. 2021{\natexlab{a}}.
\newblock \href {https://openreview.net/forum?id=zeFrfgyZln} {Approximate
  nearest neighbor negative contrastive learning for dense text retrieval}.
\newblock In \emph{Proceedings of the 9th International Conference on Learning
  Representations}. OpenReview.net.

\bibitem[{Xiong et~al.(2021{\natexlab{b}})Xiong, Wang, and
  Wang}]{DBLP:conf/eacl/XiongWW21}
Wenhan Xiong, Hong Wang, and William~Yang Wang. 2021{\natexlab{b}}.
\newblock \href {https://doi.org/10.18653/v1/2021.eacl-main.244} {Progressively
  pretrained dense corpus index for open-domain question answering}.
\newblock In \emph{Proceedings of the 16th Conference of the European Chapter
  of the Association for Computational Linguistics: Main Volume}, pages
  2803--2815, Online. Association for Computational Linguistics.

\bibitem[{Xu et~al.(2020)Xu, Hamzei, Nyamsuren, Kruiger, Winter, Tomko, and
  Scheider}]{agile-giss-1-23-2020}
H.~Xu, E.~Hamzei, E.~Nyamsuren, H.~Kruiger, S.~Winter, M.~Tomko, and
  S.~Scheider. 2020.
\newblock \href {https://doi.org/10.5194/agile-giss-1-23-2020} {Extracting
  interrogative intents and concepts from geo-analytic questions}.
\newblock \emph{AGILE: GIScience Series}, 1:23.

\bibitem[{Yang et~al.(2019)Yang, Dai, Yang, Carbonell, Salakhutdinov, and
  Le}]{XLNet}
Zhilin Yang, Zihang Dai, Yiming Yang, Jaime~G. Carbonell, Ruslan Salakhutdinov,
  and Quoc~V. Le. 2019.
\newblock \href
  {https://proceedings.neurips.cc/paper/2019/hash/dc6a7e655d7e5840e66733e9ee67cc69-Abstract.html}
  {{XLNet}: Generalized autoregressive pretraining for language understanding}.
\newblock In \emph{Advances in Neural Information Processing Systems 32: Annual
  Conference on Neural Information Processing Systems 2019}, pages 5754--5764.

\bibitem[{Yuan et~al.(2021)Yuan, Liu, Liu, Liu, Yang, Hu, and
  Xiong}]{IncreSTGL}
Zixuan Yuan, Hao Liu, Junming Liu, Yanchi Liu, Yang Yang, Renjun Hu, and Hui
  Xiong. 2021.
\newblock \href {https://doi.org/10.1145/3442381.3449810} {Incremental
  spatio-temporal graph learning for online query-poi matching}.
\newblock In \emph{Proceedings of the Web Conference 2021}, pages 1586--1597.
  {ACM} / {IW3C2}.

\bibitem[{Zhang et~al.(2020)Zhang, Wu, Zhou, Duan, Zhao, and Wang}]{SG-Net}
Zhuosheng Zhang, Yuwei Wu, Junru Zhou, Sufeng Duan, Hai Zhao, and Rui Wang.
  2020.
\newblock \href {https://aaai.org/ojs/index.php/AAAI/article/view/6511}
  {{SG-Net}: Syntax-guided machine reading comprehension}.
\newblock In \emph{Proceedings of the Thirty-Fourth {AAAI} Conference on
  Artificial Intelligence}, pages 9636--9643. {AAAI} Press.

\bibitem[{Zhao et~al.(2019{\natexlab{a}})Zhao, Peng, Wu, Chen, Yu, Zheng, Ma,
  Chai, Ye, and Qie}]{PALM}
Ji~Zhao, Dan Peng, Chuhan Wu, Huan Chen, Meiyu Yu, Wanji Zheng, Li~Ma, Hua
  Chai, Jieping Ye, and Xiaohu Qie. 2019{\natexlab{a}}.
\newblock \href {https://doi.org/10.1609/aaai.v33i01.33011270} {Incorporating
  semantic similarity with geographic correlation for query-poi relevance
  learning}.
\newblock In \emph{Proceedings of the Thirty-Third {AAAI} Conference on
  Artificial Intelligence}, pages 1270--1277. {AAAI} Press.

\bibitem[{Zhao et~al.(2019{\natexlab{b}})Zhao, Zhu, Liu, Xu, Li, Zhuang, Sheng,
  and Zhou}]{Next-POI}
Pengpeng Zhao, Haifeng Zhu, Yanchi Liu, Jiajie Xu, Zhixu Li, Fuzhen Zhuang,
  Victor~S. Sheng, and Xiaofang Zhou. 2019{\natexlab{b}}.
\newblock \href {https://doi.org/10.1609/aaai.v33i01.33015877} {Where to go
  next: {A} spatio-temporal gated network for next {POI} recommendation}.
\newblock In \emph{Proceedings of the Thirty-Third {AAAI} Conference on
  Artificial Intelligence}, pages 5877--5884. {AAAI} Press.

\end{thebibliography}
\bibliographystyle{acl_natbib}

\clearpage
\appendix

\section{POI Example}\label{sec:poi}
\tabref{tab:poi_example} shows a POI example, from which we can see that many reviews have similar semantics, making it important to choose representative reviews. In this work, we cluster sentences from reviews, and choose reviews evenly from each cluster to make up the textual input.

\begin{table*}[t!]
    \small
	\begin{center}
		\begin{tabularx}{\textwidth}{l|X}
			\toprule
			Key & Value  \\
			\midrule
			Name & Donnybrook, 35 Clinton St, New York City, NY 10002-2426 \\
			\midrule
			Lat Long & [40.7201861, -73.9846227] \\
			\midrule
			\multirow{25}{*}{Reviews} & The place was far from packed, but those who were there were very loud. \\
			& It was the only place that we didnt have on a list of places to go and I have to say it was one of the high lights of the night. \\
			& We stumbled across this place whilst enjoying the night life around the lower east side late on a Saturday night. \\
			& I went in for drinks while I waited for a reservation nearby. \\
			& On the whole , the atmosphere was not one in which I'd like to stay very long. \\
			& The loud music wasn't the problem. \\
			& Turns out that the place is very noisy. \\
			& They were not serving food when we arrived, but the bar tender ordered a pizza for us which we ate at the bar :-) Definitely include it in an East Village pub crawl \\
			& Nice bar to grab a beer and a warm pretzel. \\
			& I wanted a nice pub to sit down and have a beer in peace and quiet. \\
			& You always find a seat in this place . \\
			& A great prerequisite to the delancey for the final blow out. \\
			& A disappointment, but it depends what you're after. \\
			& We were early for our res at Ivan Ramen down the street. \\
			& Good beer and good service . \\
			& They have Magners (just what you need on a hot NYC summer day) The pace is easy going, the staff are friendly and the drinks are reasonably priced (similar to Dublin). \\
			& Staff is kindle and guinness is a real guinness . \\
			& Good draft selection and very friendly service \\
			& The bartender was very friendly. \\
			& I can't speak to the food, but it was exactly what we needed when we needed it. \\
			& I had read the reviews, and had high expectations. \\
			& For a very young audience, I guess it might be fun \\
			& The music was R\&B / HIPHOP / Pop and the whole place had a really good vibe. \\
			& Everyone up dancing, lots of new yorkers. \\
			& We stumbled across this "Irish" bar while waiting to check into our hotel. \\
			\bottomrule
		\end{tabularx}
	\end{center}
	\caption{A POI example, where reviews have been segmented into sentences.}
	\label{tab:poi_example}
\end{table*}

\begin{table*}[t!]
	\small
	\centering
	\begin{tabular}{llccccccccc}
		\toprule
        \multicolumn{2}{c}{Negatives} & \multicolumn{4}{c}{Local} && \multicolumn{4}{c}{Global} \\
          \cmidrule{3-6}
          \cmidrule{8-11}
        \#N & \#HN & Acc@3 & Acc@5 & Acc@30 & MRR && Acc@5 & Acc@30 & Acc@100 & MRR\\
		\midrule
        1* & 0  & 14.98 & 20.75 & 48.30 & 0.140 && \z3.57 & 10.64 & 21.80 & 0.020 \\
        3* & 2  & 17.22 & 23.92 & 52.95 & 0.164 && \z9.71 & 28.22 & 45.03 & 0.067 \\
        7 & 5  & 19.76 & 25.66 & 56.58 & 0.182 && 11.81 & 29.40 & 48.81 & 0.090 \\
        31 & 16 & 24.24 & 31.56 & \textbf{61.33} & 0.218 && 14.51 & 34.06 & 50.72 & 0.105 \\
        47 & 24 & \textbf{24.57} & \textbf{32.05} & 60.75 & \textbf{0.221} && \textbf{16.16} & \textbf{37.76} & \textbf{54.78} & \textbf{0.117} \\
		\bottomrule
	\end{tabular}
	\caption{Results with differing numbers of total negatives, with around 3/4 hard negatives. Lines with * signify results with early stopping, because using only hard negatives collapsed the model.}
	\label{tab:more_neg}
\end{table*}

\section{Impact of Total Number of Negative Examples}\label{sec:more_neg}
\tabref{tab:more_neg} presents experimental results with differing numbers of total negative examples.
As the training process is based on contrastive learning, increasing the number of negative examples within a batch leads to an improvement in the model's performance.

\begin{figure*}[t]
	\centering
	\includegraphics[width=1\textwidth]{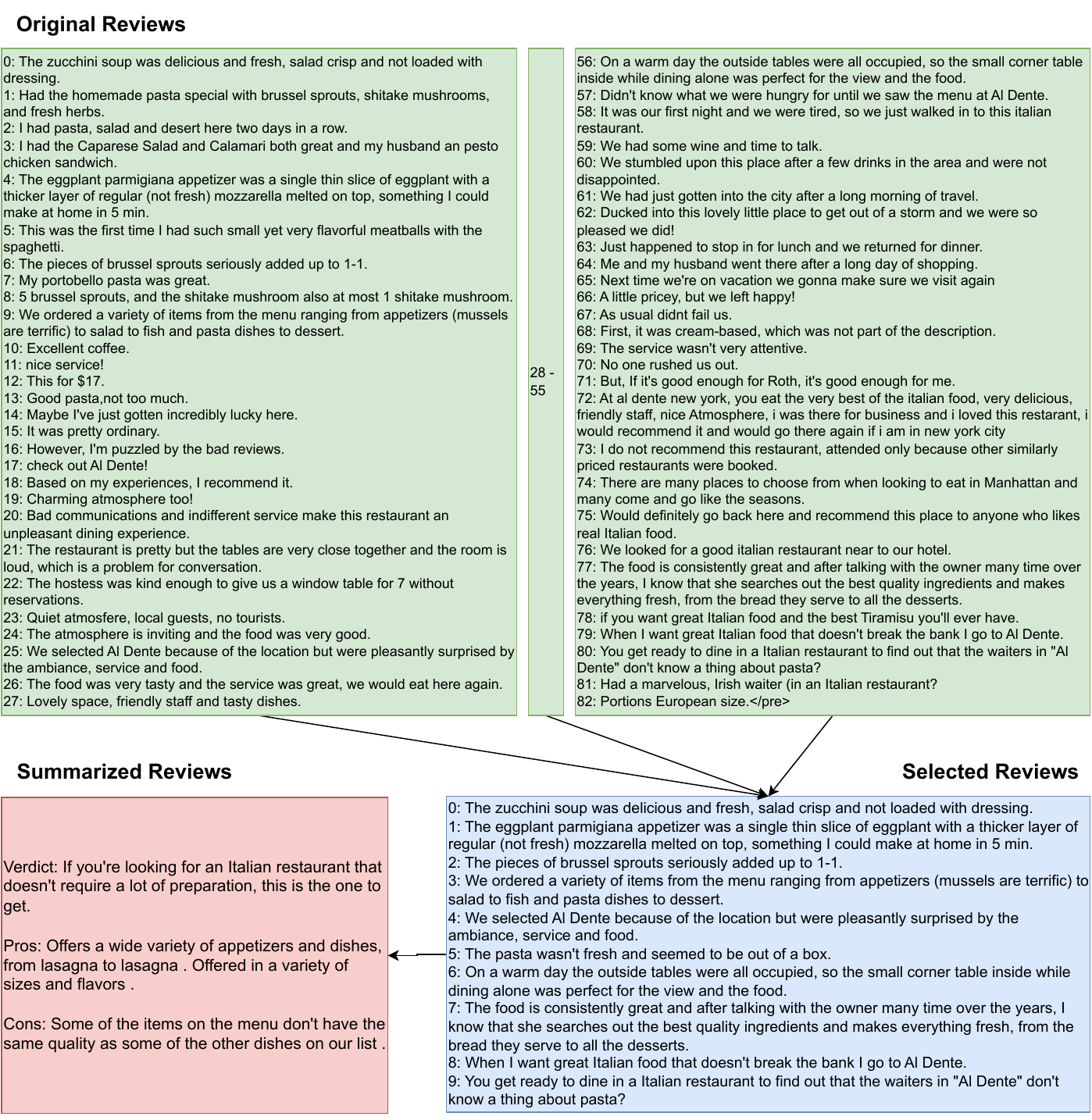}
	\caption{Example of \selsum output.}
	\label{fig:selsum_example}
\end{figure*}
\begin{table*}[h!]
	\small
	\centering
	\begin{tabular}{llccccccccc}
		\toprule
        \multirow{2}{*}{Review Module} & \multicolumn{4}{c}{Local} && \multicolumn{4}{c}{Global} \\
          \cmidrule{2-5}
          \cmidrule{7-10}
        & Acc@3 & Acc@5 & Acc@30 & MRR && Acc@5 & Acc@30 & Acc@100 & MRR\\
        \midrule
        Cluster & 23.90 & 31.20 & 60.52 & 0.216 && 13.28 & 32.12 & 48.63 & 0.096 \\ 
        \textsc{Sel} & 24.87 & 32.08 & 61.17 & 0.221 && 13.28 & 32.05 & 47.96 & 0.095 \\ 
        \selsum & 24.83 & 32.51 & 60.92 & 0.220 && 14.07 & 32.87 & 49.08 & 0.101 \\ 
		\bottomrule
	\end{tabular}
	\caption{Comparison of using clustered reviews, selected reviews with \selsum, and summarized reviews with \selsum.}
	\label{tab:selsum}
\end{table*}

\section{\selsum Example and Effectiveness}\label{sec:selsum_app}
\figref{fig:selsum_example} shows an example of \selsum model output. 
\tabref{tab:selsum} presents the comparison of using clustered reivews, selected reviews (of \selsum), and summarized reviews.

\section{Efficiency and Usability Analysis}\label{sec:usability}
The most important component of \lamb is the textual encoder, which can be replaced by any pre-trained language model.
With the increased development of model distillation and compression methods \cite{Tinybert, Minilm, Mobilebert}, \lamb can be made more space and time efficient with advanced encoders.
Here, we analyse the model's efficiency and usability by considering: (1) adding a new POI into the database; (2) answering a new question; and (3) maintaining the high accuracy of the model.

\myparagraph{New POI}: To add a new POI to the candidate set, the first step is to do inference using \selsum model.
It is then fed into the POI encoder, and stored for inference purposes.
The primary costs of GPU training time and GPU memory consumption can be ignored.

\myparagraph{New Question}: Given a new question, we input it into the question encoder without any pre-processing such as geo-parsing or tagging.
After encoding, \lamb ranks the candidate POIs according to vector similarity, based on simple vector dot product. 
In this paper, we didn't use any special techniques to speed this up, but in practical applications, techniques such as FAISS \cite{FAISS} can be used to achieve sub-linear times.\footnote{FAISS is an efficient open-source library for approximate nearest-neighbor search.}

\begin{figure}[]
	\centering
	\includegraphics[width=0.9\columnwidth]{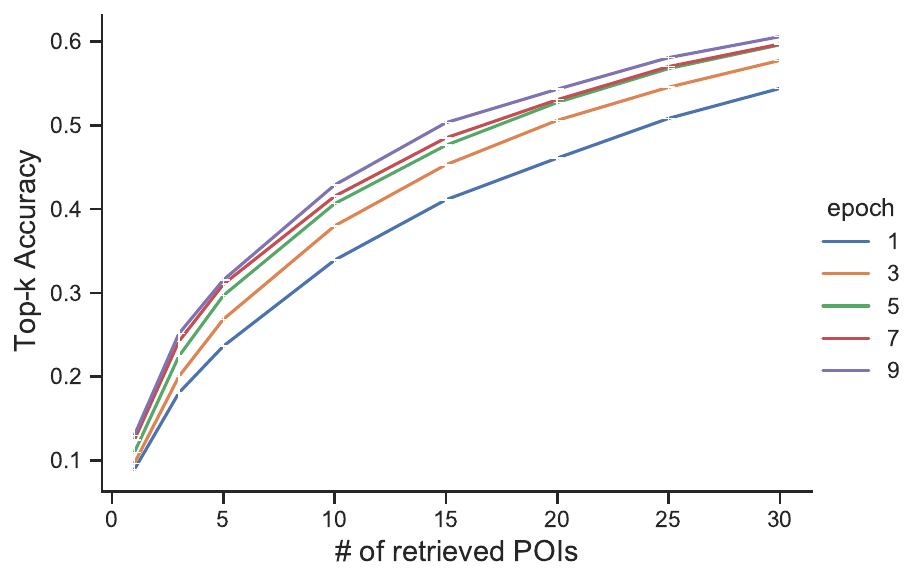}
	\caption{Top-$k$ accuracy with varying numbers of training epochs.}
	\label{fig:train_visualize}
\end{figure}

\myparagraph{Training and Update}: The training of \lamb takes no more than 12 hours on a single GPU.
\figref{fig:train_visualize} shows the top-$k$ retrieval accuracy with respect to the number of training epochs, based on which we can see that the model already achieves good results after 5 epochs.
Once this has happened, there is no need to retrain the model from scratch: as more and more new questions and POIs appear, to maintain high performance of the model, it should be enough to fine-tune it on the new questions and POIs for one or two additional epochs.

\begin{table}[t]
	\small
	\centering
	\begin{tabular}{llcccc}
		\toprule
		Module & Acc@3 & Acc@5 & Acc@30 &MRR \\
		\midrule
		\lamb Loc & \textbf{24.83} & \textbf{32.51} & \textbf{60.92} & \textbf{0.220} \\
		Geo-loc & 22.01 & 29.54 & 58.24 &  0.204\\
        Geo-dist & 20.25 & 28.00 & 58.43 & 0.189\\
		\bottomrule
	\end{tabular}
	\caption{Comparison between \lamb location module and other geo-coordinate-based location/distance modules on local evaluation.}
	\label{tab:loc_compare}
\end{table}

\section{Comparison to Geo-coordinate-based Location/Distance Module}\label{sec:distance}
We compare our location module with straightforward geo-coordinate-based location and distance modules.
Specifically, during question pre-processing, we detect location mentions and tag them with geo-coordinates using a geo-tagger.
Similar to \lamb, the question location module $E_{Q}^{loc}$ maps the geo-coordinates of the mentioned locations into fixed-length vectors:
\begin{equation*}
r_{q}^{loc} = E_{Q}^{loc}([l_1,l_2,...,l_m]) \in \mathbb{R}^{1\times d_2}
\end{equation*}
where $m$ is a hyper-parameter determined based on the average number of location mentions in questions ($m=5$ here). 
Each $l_i$ is a $2$-d vector $[lat_i, long_i]$. If a question contains $n>m$ unique locations, we randomly select $m$ locations as the input to $E_{Q}^{loc}$, otherwise we pad the input to $m$ with $[0,0]$.
Note that the output dimension $d_2$ is fixed and independent of the number of locations $n$.
For POI, we simply set $m=1$.

\paragraph{Location Module}
The location modules for both questions and POIs are implemented with a multi-layer perceptron.
Since multiple location mentions (geo-coordinates) may exist in a given question while each POI has a unique geolocation, the sizes of the two location modules are slightly different: POIs are represented as $[lat,long]$ (with size = 2), while questions are represented as $[lat_1, long_1, lat_2, long_2, ..., lat_m, long_m]$ (size = $2m$).
We use a 3-layer MLP with dropout of 0.2 and ReLU activation function to map locations into a $2m$-d vector (i.e., $d_2=2m$).

\paragraph{Distance Module}
Since the location module indiscriminately encodes location mentions from the question into a fixed-length vector, some of which may be irrelevant or even harmful for POI matching, we add a distance module to explicitly compute a distance score from the location mentions in the question to a POI, followed by min-pooling to choose the minimal distance from the question to a given POI.
We use the Haversine formula to compute distances.

To use the distance module, we define similarity between a question $q$ and a POI $p$ using the weighted sum of the bi-encoder similarity score and distance score:
\begin{align*}
\text{sim}(p,q) = (1-\lambda) \text{sim}(r_p, r_q) - \lambda (\text{dist}(p,q))
\end{align*}
We negate the distance score to ensure the closer the two locations, the higher the similarity. 
$\lambda \in [0,1]$ is a distance score weight, where  $\lambda=0$ means the model does not consider distance at all and $\lambda=1$ means the model computes scores by distance only.

We compare our location module with these straightforward geo-coordinate-based location/distance modules in \tabref{tab:loc_compare}.
From the table we can clearly see that our module is much better than the alternatives.

\end{document}